# Constrained Optimal Planning to Minimize Battery Degradation of Autonomous Mobile Robots

Jiachen Li
*Department of Mechanical Engineering*
*The University of Texas at Austin*
Austin, Texas
jiachenli@utexas.edu

Jian Chu
*Department of Mechanical Engineering*
*The University of Texas at Austin*
Austin, Texas
jian_chu@utexas.edu

Feiyang Zhao
*Department of Mechanical Engineering*
*The University of Texas at Austin*
Austin, Texas
feiyang_zhao@utexas.edu

Shihao Li
*Department of Mechanical Engineering*
*The University of Texas at Austin*
Austin, Texas
shihaoli01301@utexas.edu

Wei Li
*Department of Mechanical Engineering*
*The University of Texas at Austin*
Austin, Texas
weiwli@austin.utexas.edu

Dongmei Chen*
*Department of Mechanical Engineering*
*The University of Texas at Austin*
Austin, Texas
dmchen@me.utexas.edu
*Corresponding author

*Abstract*—This paper proposes an optimization framework that addresses both cycling degradation and calendar aging of batteries for autonomous mobile robot (AMR) to minimize battery degradation while ensuring task completion. A rectangle method of piecewise linear approximation is employed to linearize the bilinear optimization problem. We conduct a case study to validate the efficiency of the proposed framework in achieving an optimal path planning for AMRs while reducing battery aging.

## I. Introduction

Autonomous mobile robots (AMRs) have become increasingly common in industrial and commercial settings, primarily relying on batteries for power in their material handling and transportation tasks. The efficiency and longevity of these battery systems are crucial factors in reducing operational costs and maintenance expenses. Recent research has extensively focused on minimizing battery degradation while maximizing the AMR performance.

The relationship between battery degradation and charging strategies has emerged as a critical area of study [1, 2, 3], with particular attention paid to capacity fading through cycling loss. Research has shown that aggressive charging rates, while reducing downtime, can significantly accelerate battery degradation. Several key factors have been identified as primary contributors to battery degradation, including large depth of discharge [4], high state of charge (SOC) [5]. Our model does not incorporate depth of discharge (DOD). Instead, we emphasize state of charge (SOC) to study battery degradation, particularly during idle and charging periods. While DOD, which tracks how much a battery is discharged, significantly impacts cycling degradation, we focus on SOC because high levels accelerate chemical degradation through unique mechanisms.

Battery degradation occurs not only during active use but also during storage, particularly when batteries are maintained at high SOC levels without activity—a phenomenon known as calendar loss [6]. This presents unique challenges for AMR applications, where robots alternate between periods of operation and idleness. The optimal solution must balance multiple competing factors: maintaining sufficient charge for task completion, minimizing degradation through appropriate charging rates, and managing SOC levels during idle periods.

Minimizing battery degradation while maximizing the AMR performance is considered a Constrained Optimal Planning (COP) problem. The COP extends classical planning paradigms by integrating real-world limitations and requirements. The field combines mathematical optimization techniques, including dynamic programming [7] and heuristics methods to solve complex planning scenarios under multiple constraints. Also, control systems have been developed to achieve the optimal balance, prioritizing task completion while implementing strategies to extend battery life through management of charging patterns and SOC levels.

In this research, we provide a tractable algorithm for solving the COP optimization problem. While most optimal control or optimization problems are NP-hard and lack analytical solutions, traditional methods like genetic algorithms and neural networks often yield time-consuming and suboptimal results. To address this issue, the proposed framework employs advanced linearization and reformulation techniques, ensuring computational efficiency and optimality.

## II. Problem Formulation

The objective of this research is to develop an optimization framework to minimize battery degradation while completing the tasks assigned to an AMR.

### A. System Modeling

The battery system is characterized by two primary degradation mechanisms: cycling degradation and calendar aging. These mechanisms are modeled separately to capture their distinct impacts on battery life.

The cycling degradation during charging and discharging is described by [2]:

$$\eta_{cyc,i} = \left(k_1 SOC_{dev,i} \cdot e^{k_2 \cdot SOC_{avg,i}} + k_3 e^{k_4 \cdot SOC_{dev,i}}\right) \cdot Q_i \quad (1)$$

where $\eta_{cyc,i}, Q_i$ represent the degradation of battery and the capacity of the battery in one charging cycle. The parameters $k_i$ are estimated from the experimental data.

The calendar aging, which occurs during idle periods, is modeled through capacity fade [3]:

$$\frac{dQ_{\text{loss}}}{dt} = k(T, SOC) \cdot \left(1 + \frac{Q_{\text{loss}}(t)}{C_{\text{nom}}}\right)^{-\alpha(T)} \quad (2a)$$

where $k(T, SOC)$ is given by:

$$k_A \cdot e^{-\frac{E_A}{R}\left(\frac{1}{T} - \frac{1}{T_{\text{ref}}}\right)} \cdot SOC + k_B \cdot e^{-\frac{E_B}{R}\left(\frac{1}{T} - \frac{1}{T_{\text{ref}}}\right)} \quad (2b)$$

$k_A, k_B, E_A, E_B$ are experimentally determined parameters, while $R, T_{\text{ref}}$ represent the gas constant and reference temperature, respectively, and can be considered as constants.

The AMR operates within a workspace containing fixed task locations where both task execution and battery charging can occur. The robot has a workflow as shown in Fig. 1 with AMR variables. After being assigned different groups of tasks, the robot executes tasks, recharges itself to the target SOC $\bar{S}$, and waits for the next group of tasks coming. $t_i, t_{c_i}, t_{w_i}, \Delta t_i$ represent the execution time for task group $i$, charging time after task group $i$, robot idle time after task group $i$, and postponed time for task group $i$, respectively. $d_i$ represents the estimated distance for task group $i$. $\xi_i$ represents the time interval between two task group $i$ and $i+1$.

In the baseline/benchmark scenario, the target SOC is set as $\bar{S} = 0.8$, the AMR speed $v$ is set to be $\bar{v}$, and the C-rate $c$ is set to be $\bar{c}$. This configuration ensures that tasks and battery recharging are completed as efficiently as possible.

B. *Objective function*

The optimization problem aims to balance the dual goals of minimizing battery degradation and ensuring timely task completion. For a sequence of $N$ tasks, each group $i$ is assigned with time window $\xi_i$. We consider two distinct types of battery degradation that occur during operation. The first one is associated with battery charging (3), and the second one is from battery idle time (4).

$$\eta_{c_i} = f_c(c) \cdot t_{c_i} \quad (3)$$
$$\eta_{s_i} = f_s(\bar{S}) \cdot t_{w_i} \quad (4)$$
$$i = 1, 2, \dots, N$$

For completing tasks, the task waiting time $\sum \Delta t_i$ needs to be minimized. Therefore, the overall objective function is:

$$\min_{\bar{S}, v, c} \sum_{i=1}^{N} \left(\eta_{c_i} + \eta_{s_i} + \lambda \Delta t_i\right) \quad (5)$$

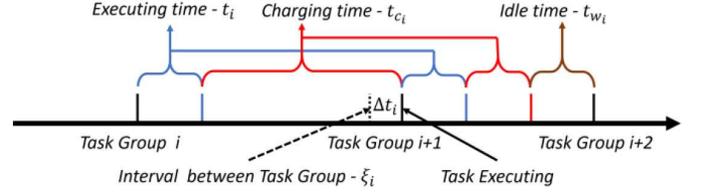

Fig. 1: Defining AMR Variables

The optimization process determines three variables: $\bar{S}, v, c$. These optimal values aim to extend battery life beyond what would be achieved with standard baseline settings while maintaining efficient task completion. The optimization problem must satisfy several constraints. First, to ensure the system can handle unexpected situations, SOC must remain above a minimum threshold $\underline{S}$. Therefore, the consumable battery capacity for executing the tasks should be lower than $\bar{S} - \underline{S}$.

$$\Delta SOC = k_v \cdot v \cdot t_i \quad (6)$$
$$\Delta SOC \leq \bar{S} - \underline{S} \quad (7)$$

where $k_v$ is the coefficient of capacity consuming rate, for simplicity, here the capacity consuming is proportional to the velocity $v$.

After executing a group of tasks, the battery needs to be recharged, therefore the charging capacity is equal to the consumed capacity

$$\Delta SOC = c \cdot t_{c_i} \quad (8)$$

The task waiting time and idle time will be calculated recursively by

$$\Delta t_{i+1} = \max\{0, \Delta t_i + t_i + t_{c_i} - \xi_i\} \quad (9)$$
$$t_{w_{i+1}} = \max\{0, \xi_i - t_i - t_{c_i} - \Delta t_i\} \quad (10)$$

Physical limits are also considered, such as robot speed limit and battery C-rate limit.

$$v \in [v^L, v^U], c \in [c^L, c^U] \quad (11)$$

The task executing time and charging time are bounded as

$$t_i \in [t^L, t^U], t_{c_i} \in [t_c^L, t_c^U] \quad (12)$$

## III. Model Linearization

The model developed in the previous section describes a battery powered AMR involving several nonlinear characteristics. The presence of bilinear terms in both the objective function and the constraints renders the optimization problem NP-hard. To address this complexity, various linearization methods are employed to approximate the nonlinear terms effectively.

The battery degradation coefficients $f_c(c)$, $f_S(\bar{S})$ are functions of C-rate $c$ and target SOC $\bar{S}$. Approximated linear relations can be derived from simulation results of (1-2), therefore, the degradation calculation can be simplified as:

$$\eta_c = k_c \cdot c \cdot t_{c_i} \quad (13)$$
$$\eta_S = k_s \cdot \bar{S} \cdot t_{w_i} \quad (14)$$

Using the equality (6) and (8), we can rewrite (13) as $\eta_c = k_c k_v d$, which is a linear term.

In (14), where the product of $\bar{S}$ and $t_{w_i}$ presents a bilinear term without any known parameters to directly facilitate its linearization, the rectangle method of piecewise linear approximation is adopted for approximation [8].

Considering $\bar{S} \in [S^L, S^U]$ and $t_{w_i} \in [t_w^L, t_w^U]$, let $w_i = \bar{S} \cdot t_{w_i}$. This approach involves dividing the variable ranges into $N_S, N_t$ intervals, $j = 1,2,\ldots,N_s, k = 1,2,\ldots,N_t$

$$S^L = S_1 < S_2 < \cdots < S_{j-1} < S_j = S^U \quad (15a)$$
$$t_w^L = t_1 < t_2 < \cdots < t_{k-1} < t_k = S^U \quad (15b)$$

For each grid cell $(k, j)$, the four corners are: $(S_j, t_k)$, $(S_{j+1}, t_k), (S_j, t_{k+1}), (S_{j+1}, t_{k+1})$. We label these four corner points accordingly as $(S_p^c, t_p^c)$ for $p = 1,2,3,4$. Within each rectangle, $w_{jk}$ is approximated using a plane equation:

$$w_{i_{jk}} = A_{jk} + B_{jk}\bar{S} + C_{jk}t_{w_i} \quad (16)$$

Then, at corner $p$, the approximation error is:

$$e_p = A_{jk} + B_{jk}S_p^c + C_{jk}t_p^c - S_p^c t_p^c \quad (17)$$

We seek $(A_{jk}, B_{jk}, C_{jk})$ that minimize the sum of squared errors:

$$\min_{A_{jk}, B_{jk}, C_{jk}} \sum_{p=1}^{4} (A_{jk} + B_{jk}S_p^c + C_{jk}t_p^c - S_p^c t_p^c)^2 \quad (18)$$

Let $X = \begin{bmatrix} 1 & S_1^c & t_1^c \\ 1 & S_2^c & t_2^c \\ 1 & S_3^c & t_3^c \\ 1 & S_4^c & t_4^c \end{bmatrix}, y = \begin{bmatrix} S_1^c t_1^c \\ S_2^c t_2^c \\ S_3^c t_3^c \\ S_4^c t_4^c \end{bmatrix}$.

We can obtain the optimal solution of (18) as [9]

$$\begin{bmatrix} A_{jk} \\ B_{jk} \\ C_{jk} \end{bmatrix}^* = (X^T X)^{-1} X^T y. \quad (19)$$

This yields the plane coefficients $(A_{jk}, B_{jk}, C_{jk})$. The plane approximation can be applied in these $N_s \cdot N_t$ pairs of intervals, and only one of these pairs can be realized at a time. Therefore, we introduce the binary variable $z_{jk} \in \{0,1\}, j = 1,2,\ldots,N_s$, $k = 1,2,\ldots,N_t$, to make sure only one constraint will be realized through the Big-M method. Here, M is chosen as a sufficiently large value to guarantee that exactly one $z_{jk}$ is 1, with all others set to 0. Also, plane approximation can reduce the approximation error by increasing the value of $N_s, N_t$.

For $j = 1,2,\ldots,N_s, k = 1,2,\ldots,N_t$, we have

$$\sum_{j=1}^{N_s} \sum_{k=1}^{N_t} z_{jk} = 1$$

$$w_{i_{jk}} \geq A_{jk} + B_{jk}\bar{S} + C_{jk}t_{w_i} - M \cdot (1 - z_{jk}) \quad (20a)$$
$$w_{i_{jk}} \leq A_{jk} + B_{jk}\bar{S} + C_{jk}t_{w_i} + M \cdot (1 - z_{jk}) \quad (20b)$$

Rewrite (20a and 20b) into the following compact matrix form:

$$e^{N \times (N_s \cdot N_t)} \cdot z^T = e^N$$

$$w_i - A - B\bar{S} - Ct_{w_i} + M \circ (e - z) \geq 0 \quad (21a)$$
$$w_i - A - B\bar{S} - Ct_{w_i} - M \circ (e - z) \leq 0 \quad (21b)$$

where $e$ denotes a vector or matrix with all ones. After linearization, the model is reformulated into a Mixed Integer Linear Programming (MILP) problem, which is tractable and could be solved by standard solvers such as MOSEK. The final model can be expressed as

$$\min_{\bar{S}, v, c} \sum_{i=1}^{N} (\eta_{c_i} + \eta_{s_i} + \lambda \Delta t_i) \quad (22)$$

$$s.t. (7), (11-14), (21)$$

## IV. Numerical Results

This section presents numerical results to demonstrate the utility of the proposed algorithm. First, we demonstrate that the linearization method achieves sufficient accuracy and efficiency. Next, we compare the results from our method and a benchmark strategy. All simulations were run on a desktop with an Intel i7-10700K CPU @ 3.8GHz, 16 GB RAM, using MATLAB 2022a and MOSEK 10.2. Here is a table summarizing the specifications for the battery and autonomous mobile robot (AMR) used in the simulation.

TABLE I. Battery and AMR Specifications

| Parameter | Value | Unit |
|---|---|---|
| Battery Type | Lithium-ion | / |
| Battery Capacity | 100 | Ah |
| State of Charge | [0, 1] | / |
| Charging Rate (C-rate) | [0.5, 2] | C |
| AMR Speed Limits | [0.5, 2] | m/s |
| Energy Consumption Rate | 0.1 | kWh/km |

### A. Accuracy and efficiency

The accuracy of our linearization method is evaluated by calculating the relative error between the actual and approximated values. We used a case study involving 5 tasks, following the predetermined trajectory illustrated in Fig. 2. We used a case study involving 5 tasks, following the predetermined trajectory illustrated in Fig. 2. This figure shows a sample route that the AMR travels through, where the battery experiences charging and depletion cycles. The red dots represent static obstacles, and the blue curve illustrates the trajectory connecting five fixed task points. The piecewise relaxation of bilinear terms depends on the selected grid size, which impacts both accuracy and computational time. To balance these factors, we choose an optimal grid density. The number of binary variables is calculated as $\#(z_{jk}) = N \cdot N_s \cdot N_t$. Since the computational time for MILP increases exponentially with the number of binary variables, a trade-off between accuracy and computational time is necessary. For typical tasks grouped in sets of three to five, the optimal parameters are $N_s = 2, N_t = 4$, resulting in high accuracy across all parameters: the errors for the task execution time approximation $t_i$ v.s. $d_i/v_i$, charging time approximation $t_c$ v.s. $\Delta SOC/c$, and idle time degradation term $\bar{S} \cdot t_w$ v.s. $w$ are 1.57%, 1.18%, and 1.04%, respectively.

To evaluate computational efficiency, we conducted a comparative test between different solution approaches. Using the genetic algorithm (GA) package in MATLAB to solve the original formulation with 10 tasks required approximately 600 seconds of computational time. In contrast, our linearized model achieves similar accuracy (with an average difference less than 1% compared to GA) while significantly reducing the computational time to approximately 30 seconds. It can be seen that the proposed algorithm significantly reduces the computational time while still achieving good accuracy.

### B. Battery SOC

The objective of optimization is to determine the $\bar{S}, c, v$ to complete a given set of tasks while minimizing battery degradation. Figure 3 compared the SOC's obtained using the proposed method and a standard rule-based algorithm [10] that is treated as a baseline standard. The proposed method achieves a lower SOC (~0.59), approximately 20% lower than the baseline. This improvement is due to the baseline, a rule-based strategy, operates at maximum speed and C-rate, leading to longer idle times and thus higher battery degradation levels. In contrast, the proposed method minimizes the SOC during idle periods by considering it in the objective function $\sum_{i=1}^{N}(\eta_{c_i} + \eta_{s_i})$.

### V. Conclusions

This paper proposes a novel COP method for AMRs to achieve the predetermined path while minimizing battery degradation. This COP framework considers battery aging, with respect to battery charging and battery idling, in the cost function for task planning. A rectangle method of piecewise linear approximation is adopted to linearize the nonlinear model. The method enables the first ever optimization framework that balances the objectives of maintaining the lowest SOC and efficiently completing the required tasks.

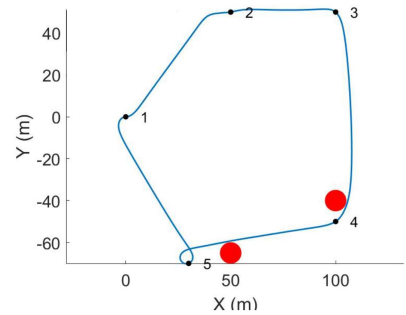

Fig. 2: AMR trajectory with Static Obstacles

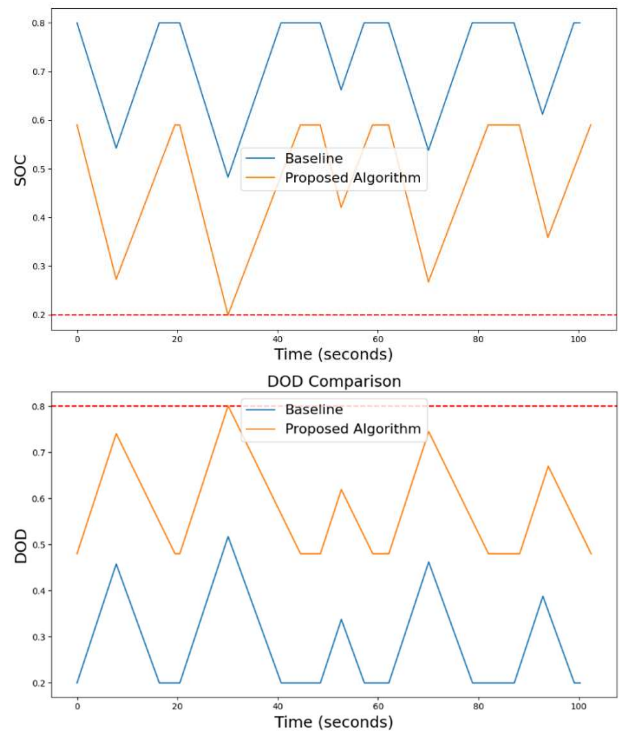

Fig. 3: Baseline & Deterministic Model Comparison in 5 Tasks Testing